\documentclass[9pt,conference]{IEEEtran}
\IEEEoverridecommandlockouts

\usepackage{cite}
\usepackage{amsmath,amssymb,amsfonts}
\usepackage{algorithmic}
\usepackage{graphicx}
\usepackage{textcomp}
\usepackage{xcolor}
\usepackage{makecell}
\usepackage{multirow}
\def\BibTeX{{\rm B\kern-.05em{\sc i\kern-.025em b}\kern-.08em
    T\kern-.1667em\lower.7ex\hbox{E}\kern-.125emX}}
\begin{document}

\title{Seek and Solve Reasoning for Table Question Answering\\
}

\author{
\IEEEauthorblockN{Ruya Jiang }
    	\IEEEauthorblockA{
		\textit{1\textsuperscript{st}Mashang Consumer Finance Co., Ltd.}\\
		Chongqing, China \\}
	    \textit{2\textsuperscript{nd}University of Toronto}\\
		\IEEEauthorblockA{
	    Toronto, Canada \\
	    ruya.jiang@mail.utoronto.ca}\\
\and
\IEEEauthorblockN{Chun Wang}
\IEEEauthorblockA{
\textit{Mashang Consumer Finance Co., Ltd.}\\
Chongqing, China \\
lukewang25@live.cn}
\and
\IEEEauthorblockN{Weihong Deng}
\IEEEauthorblockA{
\textit{Mashang Consumer Finance Co., Ltd.}\\
Chongqing, China \\
weihong.deng@msxf.com}
}

\maketitle

\begin{abstract}
The complexities of table structures and question logic make table-based question answering (TQA) tasks challenging for Large Language Models (LLMs), often requiring task simplification before solving. This paper reveals that the reasoning process during task simplification may be more valuable than the simplified tasks themselves and aims to improve TQA performance by leveraging LLMs' reasoning capabilities. We propose a Seek-and-Solve pipeline that instructs the LLM to first seek relevant information and then answer questions, integrating these two stages at the reasoning level into a coherent Seek-and-Solve Chain of Thought (SS-CoT). Additionally, we distill a single-step TQA-solving prompt from this pipeline, using demonstrations with SS-CoT paths to guide the LLM in solving complex TQA tasks under In-Context Learning settings. Our experiments show that our approaches result in improved performance and reliability while being efficient. Our findings emphasize the importance of eliciting LLMs' reasoning capabilities to handle complex TQA tasks effectively.

\end{abstract}


\section{Introduction}
\label{sec:intro}
Tables are widely used in documents, such as financial reports and government statistics, making the ability to interpret them essential. This has led to the emergence of Table-based Question Answering (TQA) as an important task. However, the complexity and diversity of table structures, such as hierarchical layouts and multi-level headers, present challenges for TQA in practice \cite{HiTab}. Additionally, natural language questions often require advanced reasoning and contextual understanding, necessitating multi-step data analysis \cite{fang2024large}. These challenges—handling complex table structures and requiring advanced reasoning skills—highlight the need for continued advancements in TQA technology to improve its practical performance and reliability.

Recently, Large Language Models (LLMs) have demonstrated strong content comprehension and reasoning abilities \cite{Mixtral}\cite{Llama3}, opening up new possibilities for addressing TQA. However, despite progress made in TQA with less complex flat tables, directly using LLMs to solve TQA with complex tables often results in less satisfactory answers \cite{TISHiTab}\cite{DATER}\cite{E5}, indicating that advanced LLMs struggle with TQA involving complex tables on their own.

Existing works \cite{DATER}\cite{E5}\cite{CABINET}\cite{ITR} typically improve complex TQA performance through task simplifications. In our terminology, these multistep approaches usually involve two major stages: the Seek stage and the Solve stage. In the Seek stage, the focus is on simplifying the complex TQA task. Common methods include identifying question-relevant data points and simplifying the task by highlighting relevant sections of the table \cite{CABINET} or creating simplified sub-tables \cite{ITR}; see Fig.~\ref{fig:tabletree} for an illustration. These operations enhance relevant information while reducing irrelevant information, making the task easier to handle. In the subsequent Solve stage, the model solves the simplified tasks instead of the raw tasks to gain benefits.

Although generally effective, the two stages are performed relatively independently. These methods primarily focus on obtaining simplified tasks, often overlooking valuable insights gained during the simplifying process that could aid in subsequent question answering. Consequently, even if the task is simpler, the Solve-stage LLM must reason from scratch. Additionally, if relevant information is erroneously omitted during simplification, the Solve-stage LLM may be unable to correct these errors \cite{CABINET}. Both factors suggest that a more holistic approach could further enhance TQA performance.

\begin{figure}[tb]
\centering
\begin{minipage}[htb]{\linewidth}
  \centering
  \centerline{\includegraphics[width=\textwidth]{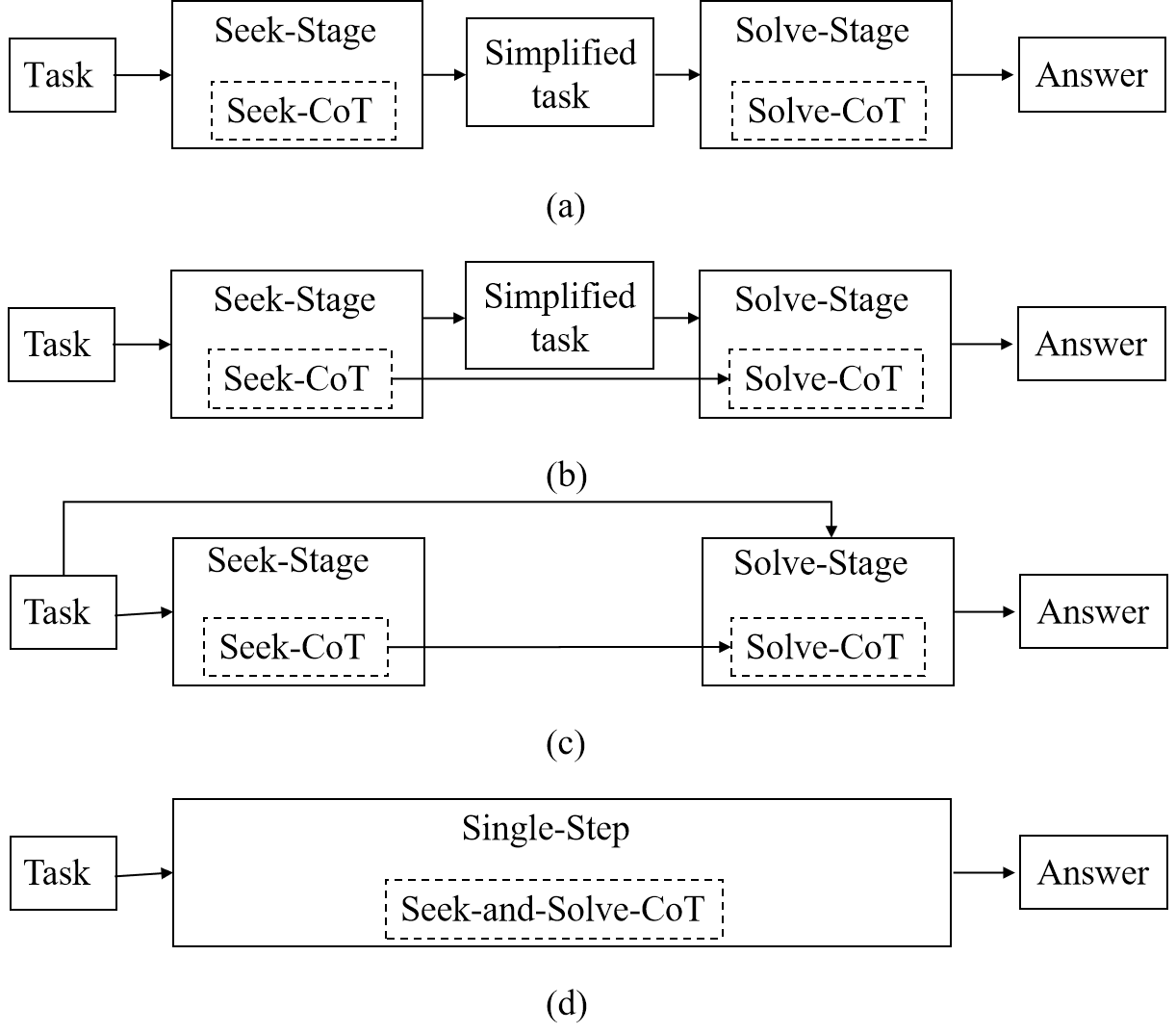}}
\end{minipage}
\caption{An illustration of different pipelines: (a) The pipeline used by existing works, characterized by simplifying the task; (b) The Seek-and-Solve pipeline, which incorporates Seek-CoT into the Solve stage in addition to generating the simplified task; (c) The Seek-and-Solve pipeline, which avoids generating the simplified task by incorporating Seek-CoT into the Solve stage to directly address the raw task; (d) The TQA-solving prompt, a compact single-step prompt.}
\label{fig:pipeline}
\end{figure}

This paper aims to improve the performance of complex TQA tasks by leveraging the reasoning capabilities of LLMs. Taking a holistic view, we first present a novel \textit{Seek-and-Solve pipeline}, which integrates the two stages at the reasoning level. In the Seek stage, the LLM is instructed to seek question-relevant information and generate both the simplified task (i.e., Seek-Result) and the Chain of Thought (CoT) \cite{COT} that leads to it (i.e., Seek-CoT). In the subsequent Solve stage, the Seek-CoT is incorporated into the Solve prompt, guiding the LLM to reason consecutively from it to solve the raw task. Extensive experiments show that Seek-CoT can be more valuable than Seek-Result. By utilizing Seek-CoT over Seek-Result, the Seek-and-Solve pipeline significantly enhances complex TQA tasks, resulting in improved performance and higher error tolerance.

Secondly, we present a compact single-step \textit{TQA-solving prompt} distilled from the pipeline, leveraging the In-Context Learning (ICL) mechanism to guide the LLM's reasoning process through demonstrations. The Seek-and-Solve-CoT (SS-CoT), representing the complete reasoning process for solving a TQA task, is formed by combining the Seek-CoT and the subsequent Solve-CoT. Experimental results show that using samples with SS-CoT paths as demonstrations, this single-step TQA-solving prompt can guide the LLM to achieve performance on par with multistep approaches. Our findings highlight that complex TQA tasks can be addressed both effectively and efficiently once the reasoning capabilities of LLMs are properly elicited.

In summary, rather than boosting TQA performance by simplifying the task, this work explores the potential to enhance TQA performance by eliciting the reasoning capabilities of the LLM. An illustrative comparison is shown in Fig.~\ref{fig:pipeline}.

The remainder of this paper is organized as follows: Sec.~\ref{sec:relatedworks} reviews related works. Sec.~\ref{sec:proposedmethods} describes the proposed methods. Sec.~\ref{sec:experiments} presents the experimental results. Finally, Sec.~\ref{sec:conclusion} concludes the paper.

\begin{figure*}[!t]
\centering
\includegraphics[width=\textwidth]{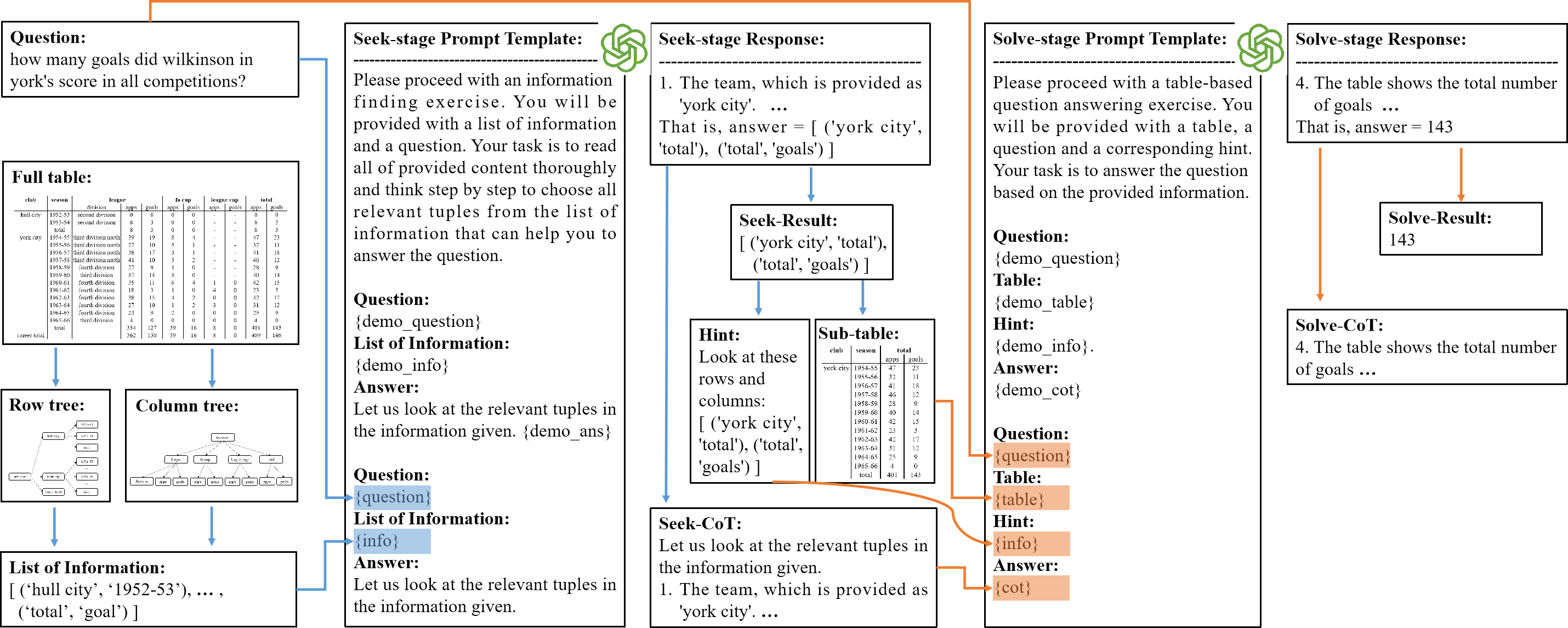}
\caption{Schematic of the Seek-and-Solve pipeline. The ICL prompts are exemplified with one-shot cases. Note, the table appears only in Solve-stage prompt.}
\label{fig:twostagepipeline}
\end{figure*}

\section{Related Works}
\label{sec:relatedworks}

\textbf{TQA with complex table}.
Due to limited token length and reasoning capabilities, TQA tasks involving complex tables often need to be simplified before answering. For example, ITR \cite{ITR} proposed creating sub-tables from relevant information identified via retrieval. E5 \cite{E5} instructed an LLM to extract relevant sections of the table. Additionally, DATER \cite{DATER} decomposed both the table and the question into sub-tables and sub-questions, respectively. Although effective, task simplification processes often operate independently of question answering (QA). Recent works tend to handle TQA tasks holistically. For example, CABINET \cite{CABINET} trained a relevance scorer with a QA language model (LM) in a differentiable manner. Additionally, \cite{LLMTableParser} instructed an LLM to perform relevant information identification and question answering in a single turn. 

Instead of focusing on obtaining the simplified task, this work explores the value of the often overlooked reasoning path that leads to it. A related study is TaCo \cite{TaCo}, which fine-tuned two LMs. The first LM generates a CoT from the table and question, and the second LM sequentially infers the answer from the CoT. Unlike TaCo, our method consists of two related tasks performed in two stages, where two-stage CoTs are performed consecutively to complete a task.

\textbf{Prompt Engineering}.
Prompt engineering is important for eliciting LLM capabilities. CoT prompting \cite{COT}, which divides reasoning into multi-step formats, has been shown to reduce errors and improve faithfulness. However, not all reasoning paths are equally effective, and some require structured guidance for optimal results \cite{stepaware}\cite{TreeOfThoughts}. ICL \cite{ICL} is another important technique that provides LLMs with demonstrations to help them solve new tasks. Combining CoT with ICL involves using demonstrations with intermediate reasoning steps to guide the model through complex problem-solving processes using structured examples \cite{COTICL}. In this work, we explore a reasoning path that incorporates domain knowledge and then combine CoT and ICL to instruct an LLM in solving TQA tasks in a dedicated manner.

\section{Proposed Methods}
\label{sec:proposedmethods}

In this section, we first introduce the tree structure as a tool to model table semantics. Next, we present the \textit{Seek-and-Solve pipeline}. Finally, we present the compact single-step \textit{TQA-solving prompt}.

\subsection{Tree structure}
\label{ssec:treestructure}
\begin{figure}[t]
\centering
\begin{minipage}[htb]{\linewidth}
  \centering
  \centerline{\includegraphics[width=\textwidth]{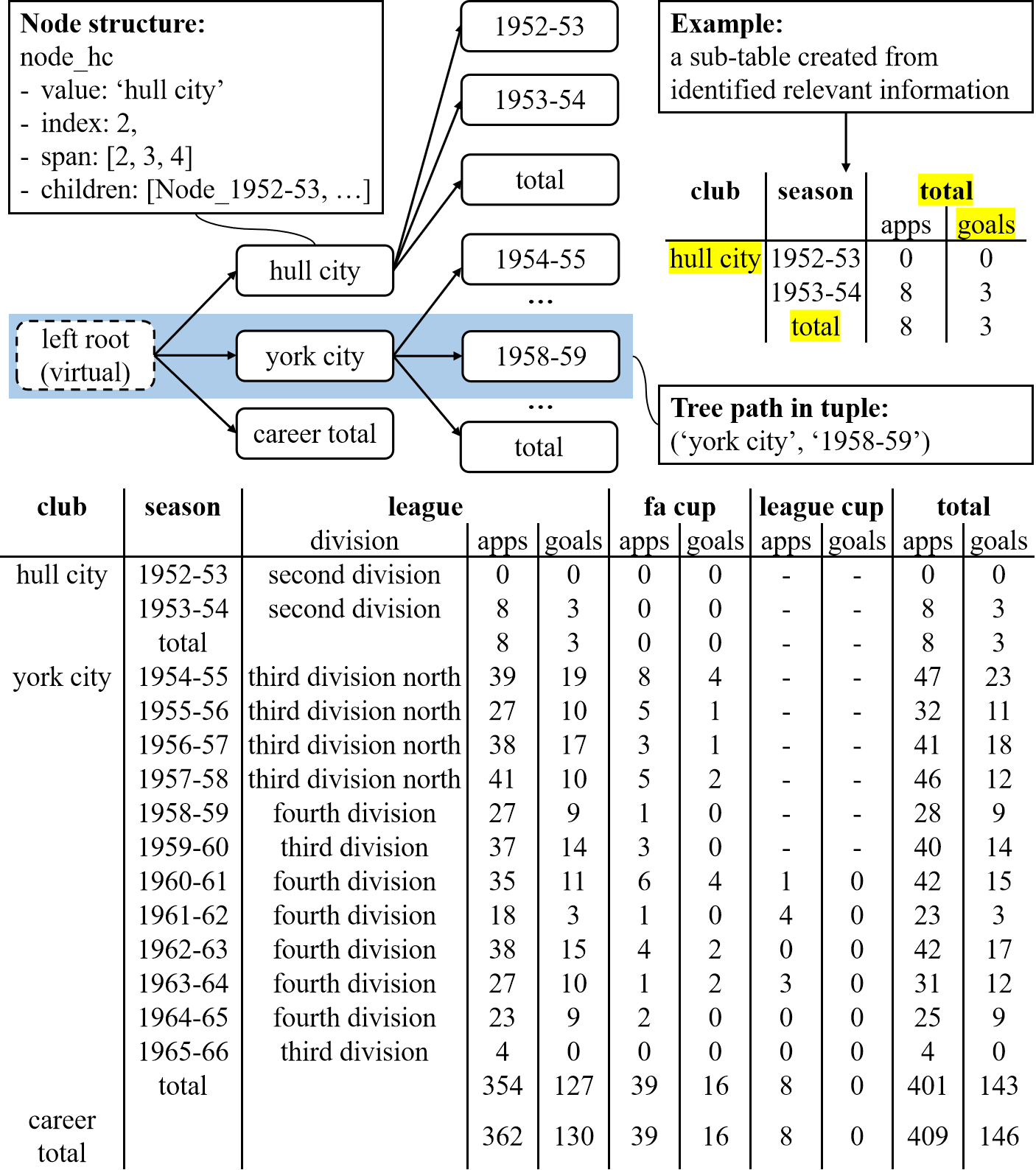}}
\end{minipage}
\caption{Illustrations of complex table and tree structure.}
\label{fig:tabletree}
\end{figure}

Previous studies \cite{LLMTableParser} \cite{multihiertt} show that modeling table header hierarchies as tree structures \cite{HiTab}\cite{TUTA} are beneficial for understanding table semantics and seeking information from complex tables. Therefore, we model the table header hierarchy as a node-based tree. In our implementation, each node corresponds to a header cell and stores its value, row (or column) index, subtree index span, and references to its child nodes. This node definition allows us to access the row (or column) index and span of any table header cell. Additionally, a tree path from the root to a leaf node can be linearized as a tuple, representing the entire hierarchical semantics of a table header. See Fig.~\ref{fig:tabletree} for an illustration. Note trees can be constructed using a model \cite{TUTA} or by parsing HTML/JSON syntax \cite{LLMTableParser}\cite{multihiertt} from a table.

\subsection{The Seek-and-Solve pipeline}
\label{ssec:twostagepipeline}
A schematic of the Seek-and-Solve pipeline is shown in Fig.~\ref{fig:twostagepipeline}. In overview, for a given TQA task, the process involves two LLM calls before arriving at an answer.

\textbf{The Seek stage.}
In this stage, we instruct the LLM to enhance its understanding of table semantics and analyze the question before performing QA by completing an information-seeking task. Specifically, we model the table's row and column header hierarchies as a row tree and a column tree, respectively. The obtained row and column trees are then converted into a list of information, where each item is a tuple linearized from a tree path. Next, we use an ICL prompt (the Seek-stage prompt template in Fig.~\ref{fig:twostagepipeline}) to instruct the LLM to analyze the question and seek question-relevant tuples (i.e., tree paths) from the list. The resulting Seek-stage response is separated into rationale and selected tuples, referred to as Seek-CoT and Seek-Result, respectively, for further use.

This work prioritizes Seek-CoT over Seek-Result, necessitating several design considerations. First, the table is excluded from the prompt to prevent the LLM from answering questions with shortcuts rather than thorough analysis. Second, we instruct the LLM to seek relevant tree paths, as reasoning steps grounded in tree paths may improve faithfulness. Third, instead of two separate lists, we combine the tree paths from both row and column trees. Empirical observations show that LLMs can misidentify a row tree path as a column tree path and vice versa. Selecting row tree paths and column tree paths in separate steps is also not recommended because it could decrease the coherence of the reasoning process. Our design addresses these concerns while encouraging the LLM to reason faithfully.

\textbf{The Solve stage.}
In this stage, we instruct the LLM to answer the question based on insights and information gained in the Seek stage.

The Seek-CoT is a comprehensive analysis that considers both table semantics and question details. In particular, its intermediate reasoning steps are linked to relevant tree paths. This ensures that each step of the reasoning process is supported by data from the table, potentially improving the overall coherence and accuracy of the question answering. Therefore, we propose allowing the LLM to reason consecutively from the Seek-CoT during the Solve stage.

In practice, the Seek-CoT can be used together with the task simplification, creating various effective combinations. We consider two common task simplification ways: providing hints \cite{CABINET} and/or subtables \cite{ITR} from the Seek-Results. As shown in Fig.~\ref{fig:twostagepipeline}, the Solve-stage prompt template consists of several slots. By filling different combinations of contents, we can form a corresponding version of the prompt.
Specifically, the following variants are considered.
\begin{itemize}
    \item \textbf{\{table\}}: This slot can hold either the full table or a sub-table derived from the Seek-Result. The sub-table is created by identifying all row and column indexes from the selected tree paths and expanding the index range by backtracking one level from the leaf node to ensure comprehensive coverage.
    \item \textbf{\{info\}}: This slot can contain either the full list of tree paths or a hint based on the Seek-Result, formatted as ``Look at these rows and columns: \{Seek-Result\}."
    \item \textbf{\{cot\}}: This slot can either involve reasoning from scratch with ``Let us think step by step" or reasoning consecutively from the Seek-CoT with ``Let us look at the relevant tuples in the information given. \{Seek-CoT\}".
\end{itemize}
Note that the contents for the slots of demonstrations change accordingly, except that demonstrations always use full tables. We use 1-shot prompt \cite{chen1shot} and Markdown formats for table representations throughout the paper. In the example shown in Fig.~\ref{fig:twostagepipeline}, sub-table, hint, and consecutive from Seek-CoT are chosen for the corresponding slots. Other prompt versions can be formed in a similar manner. 

Finally, the answer and the reasoning path, denoted as Solve-Result and Solve-CoT, can be extracted from the Solve-stage response.

\subsection{The TQA-solving prompt}
\label{ssec:qaprompt}
\begin{figure}[t]
\centering
\begin{minipage}[htb]{0.98\linewidth}
  \centering
  \centerline{\includegraphics[width=\textwidth]{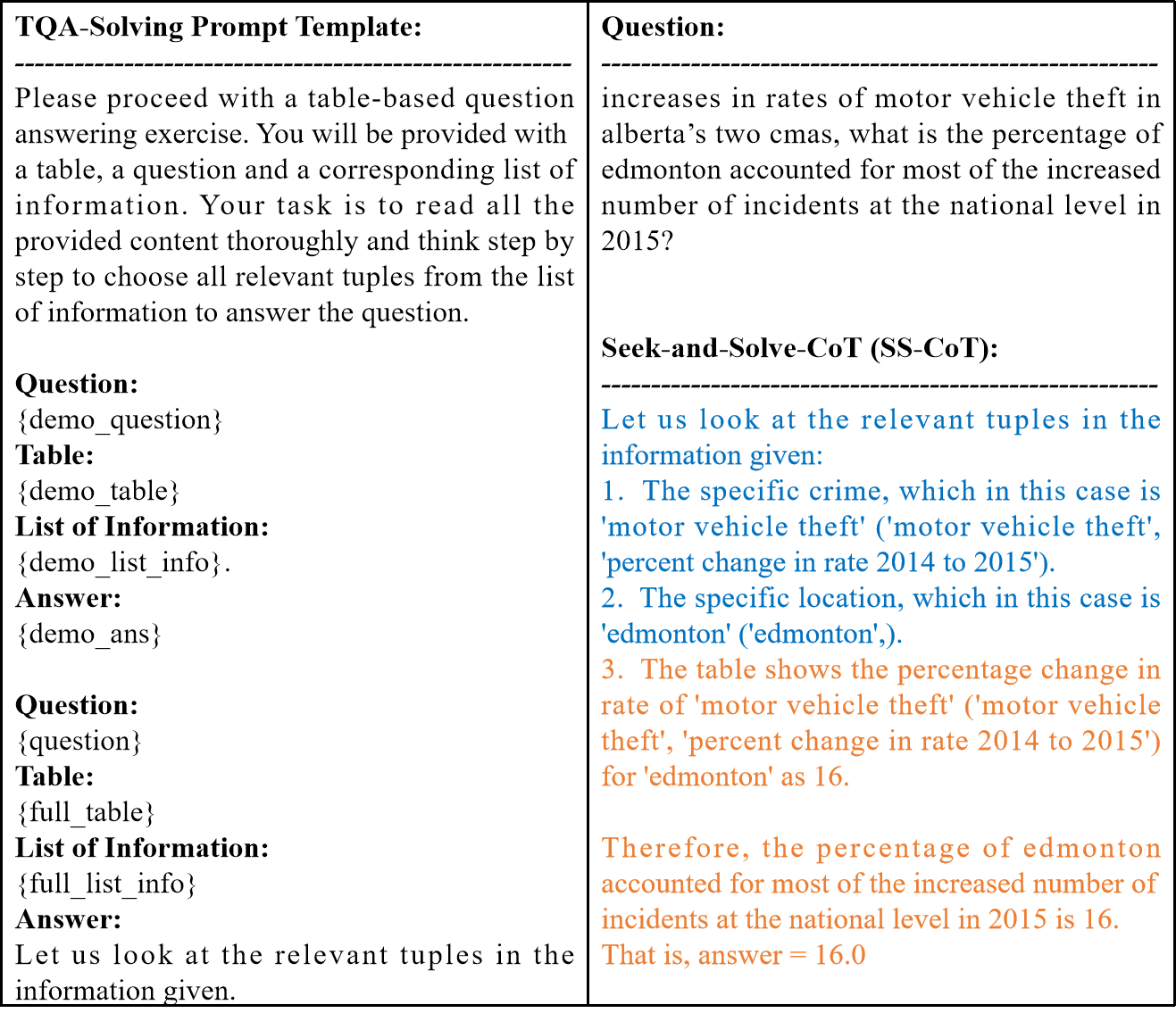}}
\end{minipage}
\caption{Illustrations of TQA-solving prompt template and SS-CoT.}
\label{fig:tqaprompt}
\end{figure}

A compact single-step TQA-solving prompt can be created by integrating the two prompts of the Seek-and-Solve pipeline, as shown in Fig.~\ref{fig:tqaprompt}. It takes the raw task, i.e., the full table and the full list of tree paths, as inputs and uses an In-Context Learning (ICL) setting to guide the LLM's reasoning process through demonstrations. Additionally, the Seek-CoT (blue) and its subsequent Solve-CoT (orange) can be integrated to form a Seek-and-Solve-CoT (SS-CoT) that represents a complete reasoning process for handling a complex TQA task, as shown in Fig.~\ref{fig:tqaprompt}. The SS-CoT distinguishes itself from reasoning paths generated natively by LLMs, referred to as vanilla-CoTs, by grounding each step on relevant table contents. 

\section{Experiments}
\label{sec:experiments}

\subsection{Experimental Settings}
\label{ssec:experimentalsettings}
\textbf{Datasets}. The proposed methods are evaluated on the HiTab \cite{HiTab} and WikiTableQuestions (WikiTQ) \cite{WikiTQ} datasets. HiTab, constructed from statistical reports and Wikipedia pages, contains diverse real-world complex tables with challenging questions. It is split into train (7,417 samples), dev (1,671 samples), and test (1,584 samples) sets. We utilize the dev and test splits for evaluations, and the train split for designing prompts. WikiTQ, known for its complex questions about Wikipedia tables, requires reasoning over multiple table data points to answer. For evaluation, we use its standard test set (4,344 samples). Following previous works \cite{HiTab} \cite{CABINET}, we use \textit{Accuracy} to measure the percentage of correctly answered samples for both datasets.

\textbf{LLMs}. The proposed methods are evaluated on four popular open-source LLMs: Mistral-7B-Instruct-v0.2 \cite{Mixtral}, Mixtral-8x7B-Instruct-v0.1 \cite{Mixtral}, Llama-3.1-8B-Instruct \cite{Llama3}, and Llama-3.1-70B-Instruct \cite{Llama3}. All models are deployed with default configurations using the vLLM library \cite{vLLM}, and accessed through APIs. In all experiments, greedy decoding and fixed random seeds are used to reduce variance.

\begin{table}[b]
\caption{The Seek-and-Solve results with different CoT on HiTab}
\begin{center}
\begin{tabular}{c|c|ccc|cc}
\Xhline{2\arrayrulewidth}
\textbf{Stage-1} &
  \textbf{Stage-2} &
  \multicolumn{3}{c|}{\textbf{Stage-2 Solve Prompt (1-shot)}} &
  \multicolumn{2}{c}{\textbf{HiTab}} \\ \hline
\textbf{LLM} &
  \textbf{LLM} &
  \multicolumn{1}{c|}{\textbf{CoT}} &
  \multicolumn{1}{c|}{\textbf{Table}} &
  \textbf{Info} &
  \multicolumn{1}{c|}{\textbf{Dev}} &
  \textbf{Test} \\ \hline
\multirow{9}{*}{\begin{tabular}[c]{@{}c@{}}Llama-\\ 3.1-8B\end{tabular}} &
  \multirow{9}{*}{\begin{tabular}[c]{@{}c@{}}Llama-\\ 3.1-8B\end{tabular}} &
  \multicolumn{1}{c|}{\multirow{5}{*}{\begin{tabular}[c]{@{}c@{}}Reason\\ from\\ scratch\end{tabular}}} &
  \multicolumn{1}{c|}{\multirow{3}{*}{\begin{tabular}[c]{@{}c@{}}Full-\\ table\end{tabular}}} &
  None &
  \multicolumn{1}{c|}{56.3} &
  59.7 \\ \cline{5-7} 
 &
   &
  \multicolumn{1}{c|}{} &
  \multicolumn{1}{c|}{} &
  Full list &
  \multicolumn{1}{c|}{54.0} &
  58.8 \\ \cline{5-7} 
 &
   &
  \multicolumn{1}{c|}{} &
  \multicolumn{1}{c|}{} &
  Hint &
  \multicolumn{1}{c|}{57.2} &
  61.2 \\ \cline{4-7} 
 &
   &
  \multicolumn{1}{c|}{} &
  \multicolumn{1}{c|}{\multirow{2}{*}{\begin{tabular}[c]{@{}c@{}}Sub-\\ table\end{tabular}}} &
  None &
  \multicolumn{1}{c|}{61.7} &
  62.6 \\ \cline{5-7} 
 &
   &
  \multicolumn{1}{c|}{} &
  \multicolumn{1}{c|}{} &
  Hint &
  \multicolumn{1}{c|}{62.8} &
  64.3 \\ \cline{3-7} 
 &
   &
  \multicolumn{1}{c|}{\multirow{4}{*}{\begin{tabular}[c]{@{}c@{}}\textbf{Consecutive} \\ \textbf{from} \\ \textbf{Seek-CoT}\end{tabular}}} &
  \multicolumn{1}{c|}{\multirow{2}{*}{\begin{tabular}[c]{@{}c@{}}Full-\\ table\end{tabular}}} &
  Full list &
  \multicolumn{1}{c|}{65.5} &
  \textbf{68.7} \\ \cline{5-7} 
 &
   &
  \multicolumn{1}{c|}{} &
  \multicolumn{1}{c|}{} &
  Hint &
  \multicolumn{1}{c|}{64.9} &
  65.9 \\ \cline{4-7} 
 &
   &
  \multicolumn{1}{c|}{} &
  \multicolumn{1}{c|}{\multirow{2}{*}{\begin{tabular}[c]{@{}c@{}}Sub-\\ table\end{tabular}}} &
  None &
  \multicolumn{1}{c|}{\textbf{67.0}} &
  67.3 \\ \cline{5-7} 
 &
   &
  \multicolumn{1}{c|}{} &
  \multicolumn{1}{c|}{} &
  Hint &
  \multicolumn{1}{c|}{66.4} &
  67.4 \\ \Xhline{2\arrayrulewidth}
\end{tabular}
\label{tab:llama31}
\end{center}
\end{table}

\begin{table}[t]
\caption{Error tolerance analysis of the pipeline with mixed LLMs}
\begin{center}
\begin{tabular}{cl|c|c|cc}
\Xhline{2\arrayrulewidth}
\multicolumn{2}{c|}{\multirow{2}{*}{\begin{tabular}[c]{@{}c@{}}\textbf{Task}\\ \textbf{Simplification}\\ (TS)\end{tabular}}} &
  \textbf{Stage-1} &
  \textbf{Stage-2} &
  \multicolumn{2}{c}{\textbf{HiTab}} \\ \cline{3-6} 
\multicolumn{2}{c|}{} &
  \textbf{LLM} &
  \textbf{LLM} &
  \multicolumn{1}{c|}{\begin{tabular}[c]{@{}c@{}}\textbf{Dev}\\ (gap)\end{tabular}} &
  \begin{tabular}[c]{@{}c@{}}\textbf{Test}\\ (gap)\end{tabular} \\ \Xhline{2\arrayrulewidth}
\multicolumn{2}{c|}{\textbf{Without} TS} &
  \begin{tabular}[c]{@{}c@{}}Llama-3.1-70B\end{tabular} &
  \multirow{3}{*}{\begin{tabular}[c]{@{}c@{}}Llama-\\ 3.1-70B\end{tabular}} &
  \multicolumn{1}{c|}{\textbf{76.7}} &
  \textbf{79.7} \\ \cline{3-3} \cline{5-6} 
\multicolumn{2}{c|}{\multirow{2}{*}{\begin{tabular}[c]{@{}c@{}}Table: Full-table\\ Info: Full-list\end{tabular}}} &
  \begin{tabular}[c]{@{}c@{}}Llama-3.1-8B\end{tabular} &
   &
  \multicolumn{1}{c|}{\begin{tabular}[c]{@{}c@{}}76.2\\ (-0.5)\end{tabular}} &
  \begin{tabular}[c]{@{}c@{}}79.0\\ (-0.7)\end{tabular} \\ \cline{3-3} \cline{5-6} 
\multicolumn{2}{c|}{} &
  \begin{tabular}[c]{@{}c@{}}Mixtral-8x7B\end{tabular} &
   &
  \multicolumn{1}{c|}{\begin{tabular}[c]{@{}c@{}}75.1\\ (-1.6)\end{tabular}} &
  \begin{tabular}[c]{@{}c@{}}78.8\\ (-0.9)\end{tabular} \\ \Xhline{2\arrayrulewidth}
\multicolumn{2}{c|}{With TS} &
  \begin{tabular}[c]{@{}c@{}}Llama-3.1-70B\end{tabular} &
  \multirow{3}{*}{\begin{tabular}[c]{@{}c@{}}Llama-\\ 3.1-70B\end{tabular}} &
  \multicolumn{1}{c|}{76.4} &
  79.6 \\ \cline{3-3} \cline{5-6} 
\multicolumn{2}{c|}{\multirow{2}{*}{\begin{tabular}[c]{@{}c@{}}Table: Full-table\\ Info: Hint\end{tabular}}} &
  \begin{tabular}[c]{@{}c@{}}Llama-3.1-8B\end{tabular} &
   &
  \multicolumn{1}{c|}{\begin{tabular}[c]{@{}c@{}}74.5\\ (-1.9)\end{tabular}} &
  \begin{tabular}[c]{@{}c@{}}77.3\\ (-2.3)\end{tabular} \\ \cline{3-3} \cline{5-6} 
\multicolumn{2}{c|}{} &
  \begin{tabular}[c]{@{}c@{}}Mixtral-8x7B\end{tabular} &
   &
  \multicolumn{1}{c|}{\begin{tabular}[c]{@{}c@{}}73.3\\ (-3.1)\end{tabular}} &
  \begin{tabular}[c]{@{}c@{}}76.1\\ (-3.5)\end{tabular} \\ \Xhline{2\arrayrulewidth}
\multicolumn{2}{c|}{With TS} &
  \begin{tabular}[c]{@{}c@{}}Llama-3.1-70B\end{tabular} &
  \multirow{3}{*}{\begin{tabular}[c]{@{}c@{}}Llama-\\ 3.1-70B\end{tabular}} &
  \multicolumn{1}{c|}{76.2} &
  77.6 \\ \cline{3-3} \cline{5-6} 
\multicolumn{2}{c|}{\multirow{2}{*}{\begin{tabular}[c]{@{}c@{}}Table: Sub-table\\ Info: Hint\end{tabular}}} &
  \begin{tabular}[c]{@{}c@{}}Llama-3.1-8B\end{tabular} &
   &
  \multicolumn{1}{c|}{\begin{tabular}[c]{@{}c@{}}72.3\\ (-3.9)\end{tabular}} &
  \begin{tabular}[c]{@{}c@{}}74.2\\ (-3.4)\end{tabular} \\ \cline{3-3} \cline{5-6} 
\multicolumn{2}{c|}{} &
  \begin{tabular}[c]{@{}c@{}}Mixtral-8x7B\end{tabular} &
   &
  \multicolumn{1}{c|}{\begin{tabular}[c]{@{}c@{}}71.6\\ (-4.6)\end{tabular}} &
  \begin{tabular}[c]{@{}c@{}}73.4\\ (-4.2)\end{tabular} \\ \Xhline{2\arrayrulewidth}
\end{tabular}
\label{tab:mixllm}
\end{center}
\end{table}

\subsection{Results with the Seek-and-Solve pipeline}
\label{ssec:resultstwostagepipeline}

Extensive experiments were conducted on HiTab to evaluate the performance of the Seek-and-Solve pipeline. As detailed in Sec.~\ref{ssec:twostagepipeline}, various prompt versions were created by combining different content elements. The results obtained using Llama-3.1-8B are presented in Tab.~\ref{tab:llama31}, with each row representing a distinct combination. Several observations can be made from these results. When the Solve-stage LLM reasoned from scratch, simplifying tasks with Seek-Results (i.e., sub-tables and/or hints) did improve results. However, when the Solve-stage LLM performed reasoning consecutively from Seek-CoT, using Seek-Results became less necessary. This is evidenced by the observation that all prompt versions consistently outperformed their counterparts by a significant margin, and the performance gaps between different prompt versions became less significant. This is likely because Seek-CoT incorporates the reasoning behind the task simplifications, largely covering the value of Seek-Results.

Furthermore, we analyzed the pipeline's error tolerance by using a more error-prone small LLM at the Seek stage and determining if a more capable Solve-stage LLM could mitigate these errors. The Seek-stage LLM could make two types of errors: one is a Seek-Result Error, caused by omitting relevant information, resulting in a subtable that fails to cover necessary areas or a hint pointing to inaccurate parts. The other is a Seek-CoT Error, which refers to errors in the Seek-stage reasoning process. Tab.~\ref{tab:mixllm} presents the results obtained using Llama-3.1-8B, Mixtral-8x7B, and Llama-3.1-70B. When certain task simplifications based on Seek-Results (i.e., sub-tables and/or hints) were applied, the more capable Solve-stage LLM struggled to correct the errors from the Seek stage, resulting in nontrivial performance gaps. This observation aligned with \cite{CABINET}. Conversely, without task simplification, errors made in the Seek stage were effectively corrected by Llama-3.1-70B in the Solve stage, as evidenced by the relatively high performance and small performance gaps. These findings revealed that the Solve-stage LLM is more effective at correcting Seek-CoT Errors than Seek-Result Errors.

Overall, by using Seek-CoT with raw tasks, the Seek-and-Solve pipeline inherently avoids Seek-Result Errors while eliciting the Solve-stage LLM to solve TQA tasks and correct Seek-CoT Errors, resulting in clear improvements in performance and error tolerance.

\begin{table}[t]
\caption{TQA-solving Prompt Results on HiTab}
\begin{center}
\begin{tabular}{ccc|cc}
\Xhline{2\arrayrulewidth}
\multicolumn{3}{c|}{\textbf{Finetuning-based Methods}} &
  \multicolumn{2}{c}{\textbf{HiTab}} \\ \hline
\multicolumn{1}{c|}{\textbf{Name}} &
  \multicolumn{1}{c|}{\textbf{Model}} &
  \textbf{\# params} &
  \multicolumn{1}{c|}{\textbf{Dev}} &
  \textbf{Test} \\ \hline
\multicolumn{1}{c|}{TAPAS \cite{HiTab}} &
  \multicolumn{1}{c|}{-} &
  - &
  \multicolumn{1}{c|}{41.2} &
  40.1 \\ \hline
\multicolumn{1}{c|}{MAPO \cite{HiTab}} &
  \multicolumn{1}{c|}{-} &
  - &
  \multicolumn{1}{c|}{44.8} &
  44.3 \\ \Xhline{2\arrayrulewidth}
\multicolumn{3}{c|}{\textbf{Prompting-based Methods}} &
  \multicolumn{2}{c}{\textbf{HiTab}} \\ \hline
\multicolumn{1}{c|}{\textbf{Name}} &
  \multicolumn{1}{c|}{\textbf{Model}} &
  \textbf{\# params} &
  \multicolumn{1}{c|}{\textbf{Dev}} &
  \textbf{Test} \\ \hline
\multicolumn{1}{c|}{Zhao et al. \cite{LLMTableParser}} &
  \multicolumn{1}{c|}{\begin{tabular}[c]{@{}c@{}}GPT-3.5 \\ (text-davinci-003)\end{tabular}} &
  - &
  \multicolumn{1}{c|}{49.0} &
  50.0 \\ \hline
\multicolumn{1}{c|}{Zero-shot CoT \cite{E5}} &
  \multicolumn{1}{c|}{\multirow{3}{*}{GPT-4}} &
  \multirow{3}{*}{-} &
  \multicolumn{1}{c|}{-} &
  78.7 \\ \cline{1-1} \cline{4-5} 
\multicolumn{1}{c|}{ReAct \cite{E5}} &
  \multicolumn{1}{c|}{} &
   &
  \multicolumn{1}{c|}{-} &
  81.9 \\ \cline{1-1} \cline{4-5} 
\multicolumn{1}{c|}{\begin{tabular}[c]{@{}c@{}}E5 \cite{E5} (with tools)\end{tabular}} &
  \multicolumn{1}{c|}{} &
   &
  \multicolumn{1}{c|}{-} &
  \textbf{85.1} \\ \Xhline{2\arrayrulewidth}
\multicolumn{1}{c|}{\multirow{4}{*}{\begin{tabular}[c]{@{}c@{}}Ours:\\ TQA prompt\\ with \textit{vanilla-CoT}\\ demonstrations\end{tabular}}} &
  \multicolumn{1}{c|}{Mistral-7B} &
  7B &
  \multicolumn{1}{c|}{45.3} &
  47.0 \\ \cline{2-5} 
\multicolumn{1}{c|}{} &
  \multicolumn{1}{c|}{Mixtral-8x7B} &
  8x7B &
  \multicolumn{1}{c|}{58.6} &
  65.3 \\ \cline{2-5} 
\multicolumn{1}{c|}{} &
  \multicolumn{1}{c|}{Llama-3.1-8B} &
  8B &
  \multicolumn{1}{c|}{58.5} &
  63.6 \\ \cline{2-5} 
\multicolumn{1}{c|}{} &
  \multicolumn{1}{c|}{Llama-3.1-70B} &
  70B &
  \multicolumn{1}{c|}{74.3} &
  78.2 \\ \Xhline{2\arrayrulewidth}
\multicolumn{1}{c|}{\multirow{4}{*}{\begin{tabular}[c]{@{}c@{}}Ours:\\ TQA prompt\\ with \textit{SS-CoT}\\ demonstrations\end{tabular}}} &
  \multicolumn{1}{c|}{Mistral-7B} &
  7B &
  \multicolumn{1}{c|}{48.7} &
  51.1 \\ \cline{2-5} 
\multicolumn{1}{c|}{} &
  \multicolumn{1}{c|}{Mixtral-8x7B} &
  8x7B &
  \multicolumn{1}{c|}{66.8} &
  71.0 \\ \cline{2-5} 
\multicolumn{1}{c|}{} &
  \multicolumn{1}{c|}{Llama-3.1-8B} &
  8B &
  \multicolumn{1}{c|}{63.0} &
  66.5 \\ \cline{2-5} 
\multicolumn{1}{c|}{} &
  \multicolumn{1}{c|}{Llama-3.1-70B} &
  70B &
  \multicolumn{1}{c|}{\textbf{76.1}} &
  79.1 \\ \Xhline{2\arrayrulewidth}
\end{tabular}
\label{tab:resultshitab}
\end{center}
\end{table}

\begin{table}[t]
\caption{TQA-solving Prompt Results on WikiTab}
\begin{center}
\begin{tabular}{ccc|c}
\Xhline{2\arrayrulewidth}
\multicolumn{3}{c|}{\textbf{Finetuning-based Methods}}                                                & \textbf{WikiTQ} \\ \hline
\multicolumn{1}{c|}{\textbf{Name}} & \multicolumn{1}{c|}{\textbf{Model}}         & \textbf{\# params} & \textbf{Test}   \\ \hline
\multicolumn{1}{c|}{TAPAS \cite{tapas}}         & \multicolumn{1}{c|}{-}                      & -                  & 48.8            \\ \hline
\multicolumn{1}{c|}{OmniTab \cite{OmniTab}}       & \multicolumn{1}{c|}{-}                      & -                  & 62.8            \\ \hline
\multicolumn{1}{c|}{CABINET \cite{CABINET}}       & \multicolumn{1}{c|}{-}                      & -                  & 69.1            \\ \Xhline{2\arrayrulewidth}
\multicolumn{3}{c|}{\textbf{Prompting-based Methods}}                                                 & \textbf{WikiTQ} \\ \hline
\multicolumn{1}{c|}{\textbf{Name}} & \multicolumn{1}{c|}{\textbf{Model}}         & \textbf{\# params} & \textbf{Test}   \\ \hline
\multicolumn{1}{c|}{Binder \cite{Binder}} &
  \multicolumn{1}{c|}{\multirow{3}{*}{\begin{tabular}[c]{@{}c@{}}GPT-3\\ (Codex)\end{tabular}}} &
  \multirow{3}{*}{175B} &
  64.6 \\ \cline{1-1} \cline{4-4} 
\multicolumn{1}{c|}{LEVER \cite{LEVER}}         & \multicolumn{1}{c|}{}                       &                    & 65.8            \\ \cline{1-1} \cline{4-4} 
\multicolumn{1}{c|}{DATER \cite{DATER}}         & \multicolumn{1}{c|}{}                       &                    & 65.9            \\ \hline
\multicolumn{1}{c|}{ReAct \cite{E5}}         & \multicolumn{1}{c|}{\multirow{2}{*}{GPT-4}} & \multirow{2}{*}{-} & 57.3            \\ \cline{1-1} \cline{4-4} 
\multicolumn{1}{c|}{E5 \cite{E5}}            & \multicolumn{1}{c|}{}                       &                    & 65.5            \\ \Xhline{2\arrayrulewidth}
\multicolumn{1}{c|}{\multirow{4}{*}{\begin{tabular}[c]{@{}c@{}}Ours:\\ TQA prompt\\ with \textit{vanilla-CoT}\\ demonstrations\end{tabular}}} &
  \multicolumn{1}{c|}{Mistral-7B} &
  7B &
  44.2 \\ \cline{2-4} 
\multicolumn{1}{c|}{}              & \multicolumn{1}{c|}{Mixtral-8x7B}           & 8x7B               & 65.4            \\ \cline{2-4} 
\multicolumn{1}{c|}{}              & \multicolumn{1}{c|}{Llama-3.1-8B}           & 8B                 & 63.7            \\ \cline{2-4} 
\multicolumn{1}{c|}{}              & \multicolumn{1}{c|}{Llama-3.1-70B}          & 70B                & 71.9            \\ \Xhline{2\arrayrulewidth}
\multicolumn{1}{c|}{\multirow{4}{*}{\begin{tabular}[c]{@{}c@{}}Ours:\\ TQA prompt\\ with \textit{SS-CoT}\\ demonstrations\end{tabular}}} &
  \multicolumn{1}{c|}{Mistral-7B} &
  7B &
  46.6 \\ \cline{2-4} 
\multicolumn{1}{c|}{}              & \multicolumn{1}{c|}{Mixtral-8x7B}           & 8x7B               & 61.9            \\ \cline{2-4} 
\multicolumn{1}{c|}{}              & \multicolumn{1}{c|}{Llama-3.1-8B}           & 8B                 & 63.0            \\ \cline{2-4} 
\multicolumn{1}{c|}{}              & \multicolumn{1}{c|}{Llama-3.1-70B}          & 70B                & \textbf{76.8}   \\ \Xhline{2\arrayrulewidth}
\end{tabular}
\label{tab:resultswikitab}
\end{center}
\end{table}

\subsection{Results with TQA-solving prompt}
\label{ssec:resultstqasolvingprompt}

We report the performance of the single-step TQA-solving prompt on the HiTab and WikiTQ datasets in Tab.~\ref{tab:resultshitab} and Tab.~\ref{tab:resultswikitab}, respectively, and compare these results with the state-of-the-art reports. In HiTab, the TQA-solving prompt with \textit{SS-CoT} demonstrations consistently outperformed those with \textit{vanilla-CoT} demonstrations across all four LLMs. In WikiTQ, likely due to the relatively flat structure of tables, the results were mixed. While there were performance declines on Mixtral-8x7B and Llama-3.1-8B, the TQA-solving prompt with \textit{SS-CoT} demonstrations showed improved performance compared to its \textit{vanilla-CoT} counterparts on Mixtral-8x7B and Llama-3.1-8B. Overall, results show that the TQA-solving prompt properly elicits the reasoning capabilities of LLMs, yielding performance comparable to multistep approaches while operating in a single step, demonstrating both effectiveness and efficiency.

\section{Conclusion}
\label{sec:conclusion}
This work enhances TQA performance by leveraging LLM's reasoning abilities. We reveal the value of the reasoning process during task simplification and propose a Seek-and-Solve pipeline that integrates two stages at the reasoning level. Additionally, we present a single-step TQA-solving prompt to guide the LLM’s reasoning process through SS-CoT demonstrations. Results demonstrate clear improvements in performance and reliability, underscoring the importance of eliciting reasoning capabilities in LLMs for complex TQA tasks. Broadly, our findings may be inspiring to tasks where reasoning in the prior stage is more valuable than its output.

\bibliographystyle{IEEEtran}
\bibliography{icassp2025}

\end{document}